\documentclass[preprint,12pt]{elsarticle}

\usepackage{amsmath}
\usepackage{amssymb}
\usepackage{amsfonts}
\usepackage{amsthm}
\usepackage{hyperref}

\journal{Journal of ABC}

\begin{document}

\begin{frontmatter}



\title{A three in one bottom-up framework for simultaneous semantic segmentation, instance segmentation and classification of multi-organ nuclei in digital cancer histology}


\author[inst1]{Ibtihaj Ahmad}

\affiliation[inst1]{organization={School of Computer Science},
            addressline={Northwestern Polytechnical University}, 
            city={Xi'an},
            postcode={710072}, 
            state={Shaanxi},
            country={PR China, }, e-mail: ibtihajahmad@mail.nwpu.edu.cn}

\author[inst2]{Syed Muhammad Israr}

\affiliation[inst2]{organization={ School of Information Science and Technology},
            addressline={University of Science and Technology of China}, 
            city={Hefei},
            postcode={230000}, 
            state={Anhui},
            country={PR China, }, e-mail: misrarustc@mail.ustc.edu.cn}

\author[inst3]{Zain Ul Islam}

\affiliation[inst3]{organization={ICT Division, College of Science \& Engineering},
            addressline={Hamad Bin Khalifa University}, 
            city={Education City},
            postcode={34110}, 
            state={Doha},
            country={Qatar, }, e-mail: zainulislam@fui.edu.pk }

\begin{abstract}
Simultaneous segmentation and classification of nuclei in digital histology play an essential role in computer-assisted cancer diagnosis; however, it remains challenging. The highest achieved binary and multi-class Panoptic Quality (PQ) remains as low as 0.68 bPQ and 0.49 mPQ, respectively. It is due to the higher staining variability, variability across the tissue, rough clinical conditions, overlapping nuclei, and nuclear class imbalance. The generic deep-learning methods usually rely on end-to-end models, which fail to address these problems associated explicitly with digital histology. In our previous work, DAN-NucNet, we resolved these issues for semantic segmentation with an end-to-end model. This work extends our previous model to simultaneous instance segmentation and classification. We introduce additional decoder heads with independent weighted losses, which produce semantic segmentation, edge proposals, and classification maps. We use the outputs from the three-head model to apply post-processing to produce the final segmentation and classification. Our multi-stage approach utilizes edge proposals and semantic segmentations compared to direct segmentation and classification strategies followed by most state-of-the-art methods. Due to this, we demonstrate a significant performance improvement in producing high-quality instance segmentation and nuclei classification. We have achieved a 0.841 Dice score for semantic segmentation, 0.713 bPQ scores for instance segmentation, and 0.633 mPQ for nuclei classification. Our proposed framework is generalized across 19 types of tissues. Furthermore, the framework is less complex compared to the state-of-the-art.
\end{abstract}


\end{frontmatter}


\section{Introduction}
\label{sec:introduction}
Digital histology plays a significant role in the analysis and diagnosis of cancers. Digital histology is in the form of a Whole Slide Image (WSI), usually of size 100,000 x 100,000 pixels. A single WSI may contain thousands of nuclei of various classes. The analysis of these nuclei provides information about the tumor. The appearance, morphological shape, and patterns act as essential markers. These markers give information about the type of cancer, grade of cancer \cite{Lu2018}, metastases \cite{Liu2017}, and survival prediction \cite{Alsubaie2018}. In current medical practices, pathologists analyze WSI using Computer Assisted Diagnosis (CAD). Although these semi-automatic CAD systems have significantly reduced time constraints, the analysis remains a tiresome job and requires an expert pathologist. An automated analysis greatly reduces the pathologist’s workload and improves the diagnosis time. 

Generally, nuclei segmentation and classification in digital histology are achieved using three different approaches, i.e., the top-down, the bottom-up, and the keypoint-based approaches \cite{Yao2021}. Top-down and keypoint-based approaches follow “detect and classify” strategies. In contrast, bottom-up approaches use “segment and classify” strategies. The drawback of the keypoint-based methods is that they achieve lower Segmentation Quality (SQ) since they only produce high-confidence instances. The top-down approaches are unsuitable for histology segmentation and classification since they perform poorly for closely packed nuclei objects. Bottom-up approaches such as \cite{Graham2019, Naylor2019, Zhou2019, Zhao2020}, produce high-resolution semantic segmentation, group the pixels in the form of instances, and finally classify the nuclei. The bottom-up approach is followed primarily due to its superior performance, specifically for digital histology. For nuclei classification, the bottom-up “segment and classify” strategy performs better than the “detect and classify” strategy \cite{gamper2020pannuke}. However, the segmentation performance of the bottom-up approaches significantly depends upon semantic segmentation. Although the bottom-up approaches are better than the rest, they still pose issues. This is due to the overlapping nucli, variation across organs, high staining variability, extreme clinical conditions, poor quality, artifacts, variable nuclear density (nuclei overlapping), variable magnification level, and nuclei class imbalance \cite{Xing2016, Graham2019, Yao2021}. Although current methods \cite{Graham2019, Naylor2019, Zhou2019,Graham2023,Swerdlow2023} have solved some of the problems; however, problems like variation in tissue, poor staining quality, variable nuclear density, and nuclei class imbalance still exist. 

\begin{figure}[!h]
\centering
\includegraphics[width=\columnwidth]{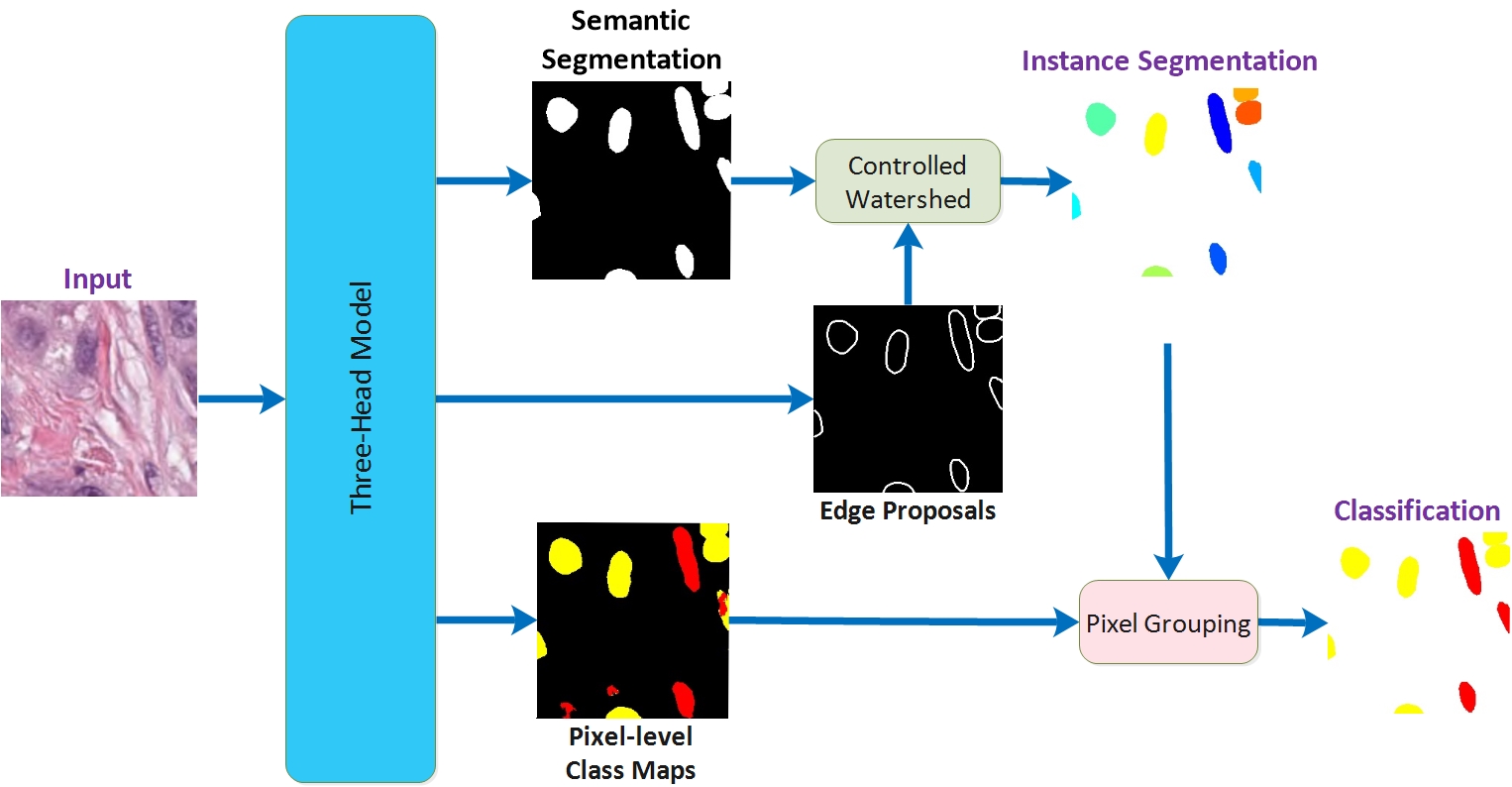}
\caption{The figure provides an overview of our bottom-up approach framework. The three-head Network produces semantic segmentation, edge proposals, and pixel-wise classes, then using post-processing, instance segmentation and classification are obtained.}
\label{fig:Main_Diagram}
\end{figure}

In our previous work, DAN-NucNet \cite{Ahmad2022}, we resolved some of these issues. However, DAN-NucNet is limited to semantic segmentation. Considering the superiority of the bottom-up approaches, we extend our previous model to a three-head dual attention model embedded in the bottom-up fashion with the post-processing methods. We extend our single decoder to a three-head decoder model. The decoder heads produce semantic segmentation, edge proposals, and pixel-level class proposals. Then we incorporate the three-head model with the post-processing methods. Our post-processing approach uses the nuclei edges to apply a controlled watershed to generate nuclei instances. These instances and the outputs from the classification decoder head are used to generate a clean version of the classified nuclei. The generic workflow of the proposed framework is shown in Fig. \ref{fig:Main_Diagram}.
The redesigned model has the following benefits: (1) Using the DAN-NucNet baseline helps improve the tissue generalization capability and the semantic segmentation quality. (2) spatial attention solves the morphological variability problem, while channel attention solves the staining variability problem. (3) Instead of direct instance segmentation, we utilize the edge proposals produced by the three-head model and the semantic segmentation. This significantly improves the instance segmentation compared to the direct approach used by the state-of-the-art. (4) The use of a different loss function for each output and the weighted loss strategy greatly improved the performance of individual output and especially the problems associated with the class imbalance. The primary contributions of this work are as follows,
\begin{enumerate}
\item{An all-in-one novel bottom-up framework for the simultaneous semantic segmentation, instance segmentation, and classification of nuclei in multi-organ cancer histology images is proposed in this work. We modify our previous model by adding additional decoder heads and post-processing methods. The baseline model is equipped with two more losses, including a weighted loss function.}
\item{We utilize the semantic segmentation and edges proposed by one of the decoder heads to perform instance segmentation, compared to the state-of-the-art, which mostly relies on direct approaches. This helps the model to predict the nuclei boundaries even in challenging conditions, such as overlapping nuclei and missing staining.}
\item{The instance segmentation and class maps are then used to assign classes to the nuclei instances. Due to this, a huge improvement in classifying imbalanced nuclear categories, i.e., connective and dead nuclei, is achieved.}
\item{The proposed framework remains less computationally extensive than the state-of-the-art. Furthermore, our framework is generalized across 19 cancer sites.}
\end{enumerate}

\section{Related Work}\label{sec:Literature_Review}
This section presents a detailed overview of the relevant literature. The related work is organized into three sub-sections based on the earlier approaches. 
\subsection{Keypoint Based Approaches}
Keypoint-based approaches regress the nuclei locality by using pre-defined key points to the nuclei. The keypoint approaches are classified into two categories, i.e., group-based and group-free key point approaches. Group-based keypoint approaches produce key points for each nucleus and then group them based on post-processing methods to form bounding boxes. CornerNet \cite{Law2018}, CentripetalNet \cite{Dong2020}, and RepPoints \cite{Yang_2019_ICCV} are examples of group-based keypoint approaches. Group-based keypoint approaches have similar drawbacks to traditional image-processing methods since they depend on traditional post-processing methods. Contrary to the group-based keypoint approaches, the group-free keypoint approaches detect the nuclei center directly. Group-free keypoint approaches include CenterNet \cite{Duan_2019_ICCV}, SaccadeNet \cite{Lan2020}, and PointNu-Net \cite{Yao2021}. The drawback of the keypoint-based methods is that they achieve lower Segmentation Quality (SQ) since they output the high confidence instances only, thus reducing the classification performance. Keypoint-based methods produce an enormous amount of incorrect bounding boxes \cite{Duan_2019_ICCV}. Furthermore, keypoint-based methods also have reduced inference speed \cite{Dong2020a}.

\subsection{Top-down Approaches}
Top-down approaches produce candidate regions and use non-maximum suppression tools to obtain bounding boxes. Then segmentation and classification of nuclei are independently performed. Examples of top-down approaches include Mask-RCNN \cite{He2017}, YOLACT \cite{Bolya_2019_ICCV}, CenterMask \cite{Lee2019}, PolarMask \cite{Xie2019}, TensorMask \cite{Chen_2019_ICCV}, and MaskLab \cite{Chen2017}. Although the top-down approaches are commonly used for generic applications, these methods are rarely used for segmentation and classification for many reasons. One reason is that these approaches have shown lower performance for segmenting nuclei boundaries due to densely populated and overlapped nuclei. These approaches may recommend tens of overlapped bounding boxes in a small region of a histology image. Furthermore, due to non-maximum suppression, a bounding box may not cover the whole nuclei.

\subsection{Bottom-up Approaches}
Bottom-up approaches produce high-resolution semantic segmentation and then utilize post-processing methods to group and clean the segmented and classified nuclei. Some top-performing bottom-up approaches include, \cite{Kumar2017}, DIST \cite{Naylor2019}, CIA-Net \cite{Zhou2019}, Triple U-Net \cite{Zhao2020}, HoVer-Net \cite{Graham2019}, and \cite{Graham2023}. Kumar et al. \cite{Kumar2017} predict the background, nuclei, and edges of the nuclei and then cluster the pixels into instances based on region growth using seeds. CIA-Net \cite{Zhou2019} uses contour supervision to generate edges. DIST \cite{Naylor2019} regresses distance maps to group pixels. HoVer-Net \cite{Graham2019} uses separate vertical and  horizontal distance maps of the nuclei pixels from their centers to achieve simultaneous instance segmentation and classification. Although these bottom-up approaches are commonly used for nuclei segmentation and classification due to their superior performance, they still possess drawbacks. These include problems like variation in tissue, poor staining quality, variable nuclear density (nuclei overlapping), and nuclei class imbalance. These methods still possess a significant performance gap in extreme clinical conditions. 

\section{Proposed Framework}\label{sec:Prop_Method}
The idea of the suggested framework is based on the bottom-up approach, i.e., we design a model that can produce basic features, which can then be utilized by the post-processing methods to produce improved segmentations and classifications. We redesign our previous model by adding additional decoder heads. These decoder heads produce edge proposals and pixel-level class proposals. It is to be noted that theoretically, we could produce instance segmentation and classifications with these additional decoders, just like some of the state-of-the-art models. However, this approach has some drawbacks. The model can easily miss-segment the overlapping and partially stained nuclei. This also results in a higher false classification rate. Therefore using additional information is necessary. In our case, we intelligently propose edges, combining them with semantic segmentation and other hand-crafted features, and then use post-processing to achieve instance segmentation and classification. Generally, our strategy is as follows. First, a histology image is given as input to our pre-trained three-head network. The three head networks produce three outputs, i.e., semantic segmentation, edge proposals, and pixel-wise classification. Next, we apply a controlled watershed algorithm using semantic segmentation and edge proposals to generate instance segmentation. Finally, we group pixels from the three-head network using instance segmentation to obtain a cleaned version of classified nuclei. The detailed working of each block is explained in the following subsections.

\subsection{Three Head Network}
\begin{figure}[!h]
\centering
\includegraphics[width=\textwidth]{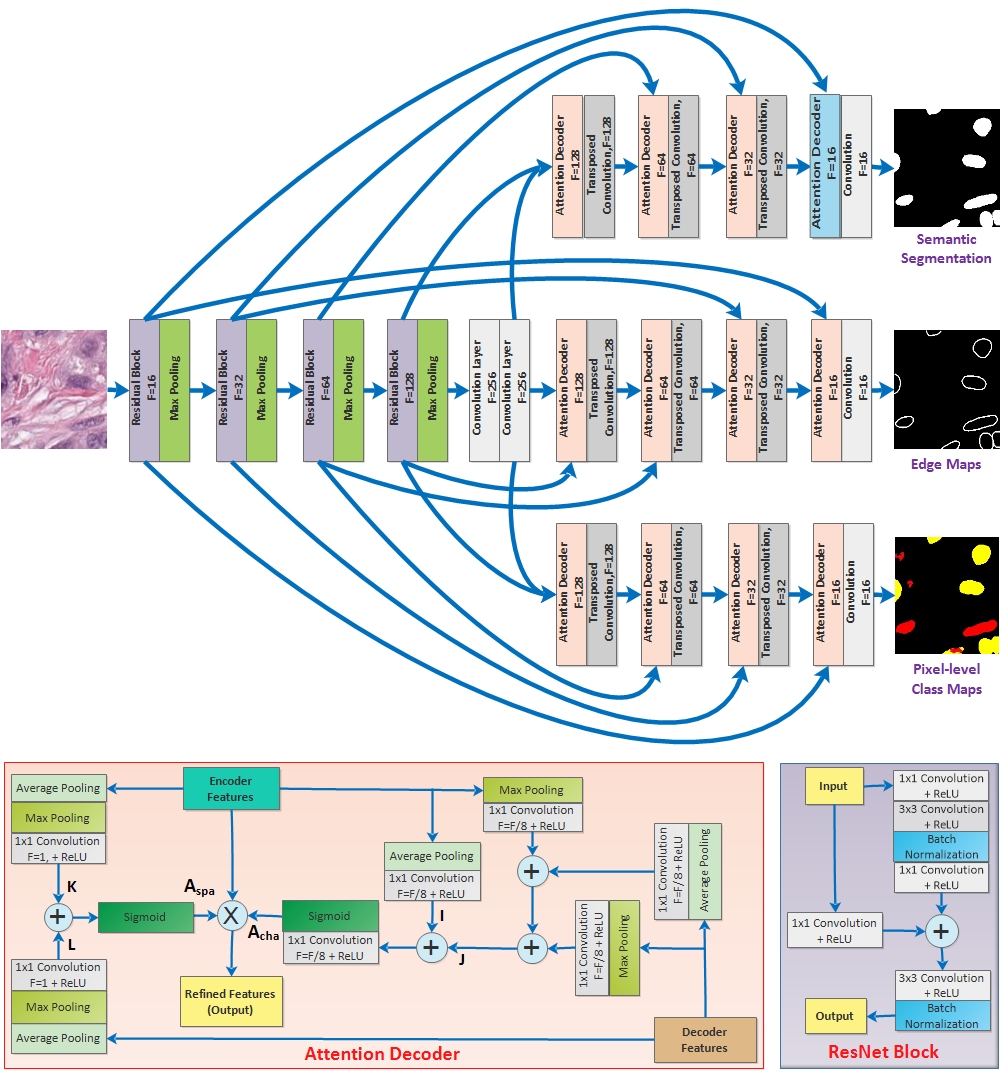}
	\caption{Overview of the three head network architecture and the detailed framework of the sub-blocks. The three head network produce three outputs, i.e., semantic segmentation, edge proposals, and pixel-wise class maps. }
	\label{fig:2_Head} 
\end{figure}

Encoder-Decoder-based networks have improved segmentation performance in digital histology compared to the rest of the networks. However, these networks possess limitations. Usually, the encoder and decoder features are directly fused. Some of the features may possess redundant information, thus increasing complexity and may even reduce performance. Digital histology is H\&E stained, where most information lies in the channel. Furthermore, some information is also spatially present; for example, the inflammatory nuclei usually surround the tumor nuclei. Therefore considering these features may increase the network performance. The second problem is the design of the encoder. The encoder is comprised of consecutive convolution layers, which produce discontinuity between lower-level and higher-level features. It also introduces the vanishing gradient problem. Inspired by these observations, we introduce spatial and channel attention mechanisms. We place our attention mechanism between the encoder and decoder as an intermediate layer. We call it the attention decoder block. Transposed convolutions follow the attention decoder blocks. We have replaced simple convolutions blocks with ResNet blocks in the encoder. Fig. \ref{fig:2_Head} shows the three-head network and building blocks in detail.
The three-head network has 16 filters at the first stage (represented by `F` in Fig. \ref{fig:2_Head}), and then at each subsequent stage of the encoder, the number of filters is doubled. At each subsequent decoder stage, the number of filters is halved.

\subsubsection{Attention Mechanism}
With some modifications, we deploy the spatial-channel attention as our previous work. We have placed transposed convolution layer at the end of the each decoder instead of placing a naive up-sample layer enabling the network to learn the features during the up-sampling. This increases performance for poorly stained images and incomplete/missing nuclei. The results of the transposed convolution are used in the successive attention decoders, except for the first decoder, where we use the output of the intermediate convolution layers. The attention mechanism is mathematically expressed as Eq. \ref{SPA_Eq}, and Eq. \ref{CHA_Eq}, respectively,

{\begin{equation} \label{SPA_Eq}
\begin{matrix}
   I = \sigma_{ReLU} \left ( C_{1,1}^{F/8} \left ( AvgPool \left ( F_{enc} \right ) \right ) \right ) \oplus \sigma_{ReLU} \left ( C_{1,1}^{F/8} \left ( MaxPool \left ( F_{enc} \right ) \right ) \right ) \\
   J = \sigma_{ReLU} \left ( C_{1,1}^{F/8} \left ( AvgPool \left ( F_{dec} \right ) \right ) \right ) \oplus \sigma_{ReLU} \left (  C_{1,1}^{F/8} \left ( MaxPool \left ( F_{dec} \right ) \right ) \right ) \\
   A_{cha} = \sigma_{sig} \left ( \sigma_{ReLU} \left ( C_{1,1}^{F/8} \left ( I \oplus J \right ) \right ) \right )
\end{matrix}
\end{equation}} 
{\begin{equation} \label{CHA_Eq}
\begin{matrix}
   K =  \sigma_{ReLU} \left (  C_{1,1}^{1} \left ( MaxPool \left ( AvgPool \left ( F_{enc} \right ) \right ) \right ) \right ) \\
   L =  \sigma_{ReLU} \left (  C_{1,1}^{1} \left ( MaxPool \left ( AvgPool \left ( F_{dec} \right ) \right ) \right ) \right ) \\
   A_{spa} = \sigma \left ( K \oplus L \right )
\end{matrix}
\end{equation}}
In the above equations $C_{k,k}^{f}$, $k$, $F$, $\sigma$, and $\oplus$ represent convolution, kernels, filters, sigmoid activation function, and concatenation, respectively. $A_{cha}$, and $A_{spa}$ represent the channel attention and spatial attention blocks output, respectively. 

The spatial attention maps $A_{spa}$ and channel attention maps $A_{cha}$ obtained above are finally multiplied with the encoder features $F_{e}$, which results in the refined features $F_{r}$. $F_{r}$ is mathematically expressed as, Eq. \ref{Attention_Fusion_Eq},
{\begin{equation} \label{Attention_Fusion_Eq}
F_{r} = F_{enc} \otimes A_{cha} \otimes A_{spa}
\end{equation}} 

\subsubsection{ResNet Block}
The proposed resnet block extracts useful semantic features at all encoder levels. The working of the ResNet block is represented in Fig. \ref{fig:2_Head} and explained as follows; First, the input to the ResNet block is passed through a 1x1 convolution + ReLU layer, then 3x3 convolution + ReLU layer, finally 1x1 convolution + ReLU layer. This is represented in Eq. \ref{ResNet_Eq1}. Simultaneously, the input is also passed through a 1x1 convolution layer, as represented in Eq. \ref{ResNet_Eq2}. The output of these two is fused, and a 3x3 convolution + ReLU layer + batch normalization is applied, which is represented as Eq. \ref{ResNet_Eq3}.

{\begin{equation} \label{ResNet_Eq1}
\left.\begin{matrix}
   T_{i,0} = \sigma_{ReLU} \left ( C_{1,1}^{F}\left ( in_{i} \right ) \right ) \\ 
   T_{i,1} = BN \left ( \sigma_{ReLU} \left ( C_{3,3}^{F}\left ( T_{i,0} \right ) \right ) \right )  \\
   X_{i,1} = \sigma_{ReLU} \left ( C_{1,1}^{F}\left ( T_{i,1} \right ) \right )  \\
\end{matrix}\right\}
\end{equation}} 

{\begin{equation} \label{ResNet_Eq2}
   X_{i,0} = \sigma_{ReLU}  \left (  C_{1,1}^{F}\left ( in_{i} \right )  \right ) 
\end{equation}} 

{\begin{equation} \label{ResNet_Eq3}
   out_{i} = BN \left ( \sigma_{ReLU} \left ( C_{3,3}^{F}\left \{ T_{i,0} \oplus T_{i.1} \right \} \right ) \right ) 
\end{equation}} 
in the above equation, $in_{i}$ represents input and $out_{i}$ represents output of the resnet block at layer $i$. $T_{i,j}$ represents intermediate features, $C_{k,k}^{F}$ represents convolution, and $BN$ represents batch normalization..

\subsection{Instance Segmentation and Classification}
\begin{figure}[!h]
\centering
\includegraphics[width=\textwidth]{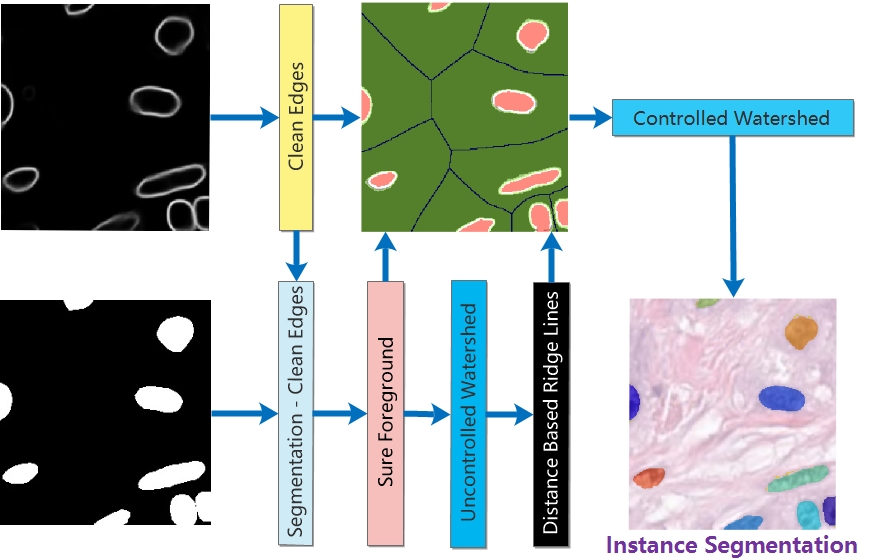}
	\caption{Illustration of the proposed instance segmentation framework.}
	\label{fig:Controlled_Watershed} 
\end{figure}

We use a controlled watershed algorithm to get nuclei instance segmentation. The instance segmentation framework is shown in Fig. \ref{fig:Controlled_Watershed}. First, we remove the edge proposals having less than ten percent probability. This limit is chosen such that the framework provides optimal performance. This step removes some of the false edges. We subtract these edge proposals from semantic segmentation maps in the next step. This separates the connected nuclei and gives us the sure foreground region (foreground markers) of the nuclei. The background markers in the semantic segmentation image are shown in black color. Ideally, we do not want these background markers to be too close to the edges of the nuclei. This is done by computing the watershed transform of the Euclidean distance transform of sure foreground markers and then identify for the watershed ridgelines. We combine all these, i.e., sure foreground, ridgelines, and background. Finally, we apply a marker-controlled watershed algorithm to get instance nuclei.

\begin{figure}[!h]
\centering
\includegraphics[width=\columnwidth]{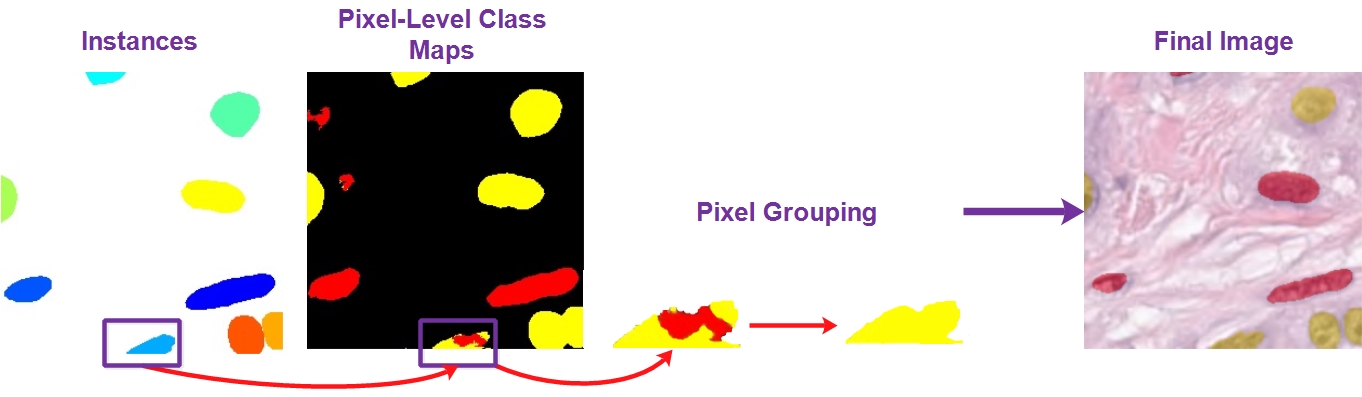}
	\caption{Illustration of pixel grouping based on instance segmentation and pixel-level class maps.}
	\label{fig:Pixel_grouping} 
\end{figure}

We take the previously segmented instances for classification and extract the corresponding region in the pixel-wise class image obtained from the three-head network. Then we separately count the pixels of each class within the extracted region. Finally we assign the selected instance to the region's class with the highest frequency. Mathematically, the process is represented as Eq. \ref{Classification_Eq}.

\begin{equation} \label{Classification_Eq}
y_{c}^{ins} =  max\left (\sum_{}^{}x_{c1}\left ( i,j \right ), \sum_{}^{}x_{c2} \left ( i,j \right ), \cdots  \sum_{}^{}x_{c3}\left ( i,j \right ) \right )
\end{equation}

where, $y_{c}^{ins}$ is the class ($c$) assigned to the instance ($ins$), $x_{cn}\left ( i,j \right )$ pixel class at row $i$, column $j$ predicted by the three head network, and $n$ is the total number of classes. An illustration of the pixel grouping is shown in Fig. \ref{fig:Pixel_grouping}. 

\subsection{Loss Function}
The proposed three head framework have three sets of losses ($l_a$, $l_b$, and $l_c$) and their respective weights ($\lambda_a$, $\lambda_b$, and $\lambda_c$) , combined under \ref{Loss_Fcn_Eq},

\begin{equation} \label{Loss_Fcn_Eq}
l = \underbrace{\lambda_a l_a}_\text{Semantic Segmentation} + \underbrace{\lambda_b l_b}_\text{Edge Proposal} + \underbrace{\lambda_c l_c}_\text{Pixel-Wise Class}
\end{equation}

For semantic segmentation and edge proposal, we retain our previous loss function \cite{Ahmad2023a, Ahmad2023}, since it has proven to be the most productive for semantic segmentation. This loss function is represented by \ref{Loss}, 

\begin{equation} \label{Loss}
l_a = \frac{Dice \; Loss * Jaccard \; Loss}{Dice \; Loss + Jaccard \; Loss} 
\end{equation}

We have used this specific loss function for semantic segmentation and edge proposals since it has shown slightly improved performance compared to the other widely used loss functions. For the classification head, we have used categorical cross-entropy loss, represented by \ref{CCE_Loss_Fcn_Eq},

\begin{equation} \label{CCE_Loss_Fcn_Eq}
l_c = - w_d \sum_{d}^{}t_{c,d}log\left ( p_{c,d} \right )
\end{equation}

where, $p$ are the predictions, $t$ are the targets, $c$ denotes the data point, $d$ denotes the nuclei class, and $w_d$ denotes individual class weights. It is to be noted that we have used separate weights ($w_d$) within the categorical cross-entropy as well in order to tackle the high nuclear category imbalance. These weights are assigned based on the percentage of the nuclear category, reported by PanNuke \cite{gamper2020pannuke}. The other weights, i.e., $\lambda_a$, $\lambda_b$, and $\lambda_c$ are, specifically, set to 1, 5, and 4, respectively, for optimal performance. The improvement in performance due to the addition of these weights is verified and reported in the ablation study (Section \ref{Sec:Ablation_studies_section}) of this work. 

\section{Experiments and Results}\label{sec:Experiments}
\subsection{Datasets}
PanNuke is the largest publically available cross-tissue nuclei cancer histology dataset with thousands of images. PanNuke dataset is obtained from combining four different datasets, i.e. CPM2017 \cite{Vu2018}, Kumar \cite{Kumar2017}, visual field from TCGA \cite{Liu2018a}, and bone marrow \cite{Kainz2015}. The images of size 256 ×256 are extracted from more than 20,000 Whole Slide Images (WSI). It contains a total of around forty-seven thousand labeled nuclei. The nuclei are labeled as instances and five classes, i.e., neoplastic, non-neoplastic or epithelial, inflammatory, connective, and dead nuclei. The dataset is split into training, validation, and testing split. We follow the same splits as followed by earlier literature in order to get reproducible results. We use limited pre-processing on the data. We removed a few blank images present in the dataset. We extract the canny edges from the labeled instances to use them for training our network head which predicts the edges of the nuclei.

\subsection{Implementation Details}
We developed the proposed network using Keras, an open-source python library for deep learning. 
The model is trained via Nvidia RTX 2060 graphical processing unit (GPU), accelerated by CUDA 11.0. We minimize the losses using the Adam optimizer with the mini-batch size of five. The learning rate is set to 0.004 at the start of the training. When the model stops learning for five consecutive epochs, the learning rate is decayed by a factor of 0.3. The training is considered as completed if the validation loss stop improving for fifteen successive epochs. 

\subsection{Results And Comparison}
We utilize PQ for comparative analysis as discussed by Graham \textit{et al.} \cite{Graham2019}. We use binary PQ (bPQ), and multi-class PQ (mPQ) as suggested by Gamper \textit{et al.} \cite{gamper2020pannuke}. bPQ assumes that all the nuclei are associated with only a single class, while mPQ calculates PQ for an individual class of nuclei and then finds the average of the individual PQ. One of the reasons for using mPQ is that it is not affected by class imbalance. To know, specifically, how our method performs over each tissue individually, we calculate bPQ, and mPQ for each of the 19 issues in the dataset individually. We report tissue-wise results and the average across the tissues in Table \ref{Result_Cmpr}. Table \ref{Result_Cmpr} also show the comparison of the suggested framework against top-performing methods i.e. HoVer-Net \cite{Graham2019}, DIST \cite{Naylor2019}, Micro-Net \cite{Raza2019}, Mask-RCNN \cite{He2017}, and PointNu-Net \cite{Yao2021}. 
\begin{table}[h]
	\caption{The table reports tissue-wise instance segmentation and classification results compared with the top-performing methods in the literature. The last row shows the average across the tissue splits. }
	\label{Result_Cmpr}
	\centering
	\tiny
	\renewcommand{\arraystretch}{1.2}
	\begin{tabular}{p{45pt}p{14pt}p{14pt}p{14pt}p{14pt}p{14pt}p{14pt}p{14pt}p{14pt}p{14pt}p{12pt}p{16pt}p{19pt}}
	\hline   
\textbf{Methods $\rightarrow$ } & \multicolumn{2}{c}{\textbf{DIST}} & \multicolumn{2}{c}{\textbf{Micro-Net}} & \multicolumn{2}{c}{\textbf{Mask-RCNN}} & \multicolumn{2}{c}{\textbf{HoVer-Net}}	 & \multicolumn{2}{c}{\textbf{PointNu-Net}}    & \multicolumn{2}{c}{\textbf{Proposed}}				\\ \hline  
\textbf{Organs $\downarrow$ }  &mPQ 	&bPQ     &mPQ 	&bPQ         &mPQ 	&bPQ	&mPQ 	      &bPQ     	&mPQ 	             &bPQ			&mPQ 	                   &bPQ 	\\   
	\hline		
AG			 	& 0.3442 & 0.5603 & 0.4153 & 0.6440	& 0.3470 & 0.5546 		& 0.4812 & 0.6962 & 0.5115 & \textbf{0.7134} & \textbf{0.6386} & 0.7031 \\
Bile Duct       	& 0.3614 & 0.5384 & 0.4124 & 0.6232	& 0.3536 & 0.5567 		& 0.4714 & 0.6696 & 0.4868 & 0.6814 & \textbf{0.6477} & \textbf{0.7557} \\
Breast          		& 0.3790 & 0.5466 & 0.4407 & 0.6029	& 0.3882 & 0.5574 		& 0.4902 & 0.6470 & 0.5147 & 0.6709 & \textbf{0.6472} & \textbf{0.7325} \\
Bladder         	& 0.4463 & 0.5625 & 0.5357 & 0.6488	& 0.5065 & 0.6049 		& 0.5792 & 0.7031 & 0.6065 & 0.7226 & \textbf{0.6654} & \textbf{0.7548} \\
Colon           	& 0.2989 & 0.4508 & 0.3414 & 0.4972 & 0.3122 & 0.4603 		& 0.4095 & 0.5575 & 0.4509 & 0.5945 & \textbf{0.6402} & \textbf{0.7164} \\
Cervix          		& 0.3371 & 0.5309 & 0.3795 & 0.6101 & 0.3402 & 0.5483 		& 0.4438 & 0.6652 & 0.5014 & 0.6899 & \textbf{0.6627} & \textbf{0.6917} \\
Esophagus       	& 0.3942 & 0.5295 & 0.4668 & 0.6011 & 0.4311 & 0.5691 		& 0.5085 & 0.6427 & 0.5504 & 0.6766 & \textbf{0.6426} & \textbf{0.7179} \\
H \& N  			& 0.3177 & 0.4764 & 0.3668 & 0.5242 & 0.3946 & 0.5457 		& 0.4530 & 0.6331 & 0.4838 & 0.6546 & \textbf{0.6229} & \textbf{0.6913} \\
Lung            		& 0.2809 & 0.4978 & 0.3370 & 0.5588 & 0.3182 & 0.5134 		& 0.4004 & 0.6302 & 0.4048 & 0.6352 & \textbf{0.6272} & \textbf{0.6977} \\
Kidney          	& 0.3339 & 0.5727 & 0.4165 & 0.6321 & 0.3553 & 0.5092 		& 0.4424 & 0.6836 & 0.5066 & 0.6912 & \textbf{0.6147} & \textbf{0.7074} \\
Liver           		& 0.3441 & 0.5818 & 0.4365 & 0.6666 & 0.4103 & 0.6085 		& 0.4974 & 0.7248 & 0.5174 & \textbf{0.7314} & \textbf{0.6050} & 0.7027 \\
Ovarian         	& 0.3789 & 0.5289 & 0.4387 & 0.6013 & 0.4337 & 0.5784 		& 0.4863 & 0.6309 & 0.5484 & \textbf{0.6863} & \textbf{0.6029} & 0.6859 \\
Prostate        	& 0.3810 & 0.5442 & 0.4341 & 0.6049 & 0.3959 & 0.5789 		& 0.5101 & 0.6615 & 0.5127 & \textbf{0.6854} & \textbf{0.6507} & 0.6821 \\
Pancreatic      	& 0.3395 & 0.5343 & 0.4041 & 0.6074 & 0.3624 & 0.5460 		& 0.4600 & 0.6491 & 0.4804 & 0.6791 & \textbf{0.6382} & \textbf{0.7246} \\
Stomach         	& 0.3369 & 0.5553 & 0.3872 & 0.6293 & 0.3684 & 0.5976 		& 0.4726 & 0.6886 & 0.4517 & 0.7010 & \textbf{0.6235} & \textbf{0.7399} \\
Skin            		& 0.2627 & 0.5080 & 0.3223 & 0.5817 & 0.2665 & 0.5021 		& 0.3429 & 0.6234 & 0.4011 & 0.6494 & \textbf{0.6309} & \textbf{0.6895} \\
Thyroid         	& 0.2574 & 0.5596 & 0.3712 & 0.6555 & 0.3037 & 0.5712 		& 0.4315 & 0.6983 & 0.4508 & \textbf{0.7076} & \textbf{0.6202} & 0.6995 \\
Testis          		& 0.3278 & 0.5548 & 0.4088 & 0.6300 & 0.3512 & 0.5420 		& 0.4754 & 0.6890 & 0.5334 & 0.7058 & \textbf{0.6666} & \textbf{0.7512} \\
Uterus          	& 0.3487 & 0.5246 & 0.3965 & 0.5821 & 0.3683 & 0.5589 		& 0.4393 & 0.6393 & 0.4846 & 0.6634 & \textbf{0.5926} & \textbf{0.7139} \\    \hline  
Average			& 0.3406 & 0.5346 & 0.4059 & 0.6053 & 0.3688 & 0.5528 		& 0.4629 & 0.6596 & 0.4957 & 0.6808 &\textbf{0.6337} &\textbf{0.7135}	\\ \hline
	\end{tabular}
\end{table}
The suggested framework achieves state-of-the-art performance compared to the top performers both for instance segmentation and classification (refer to Table \ref{Result_Cmpr}). We report average PQ for each nuclei class in Table \ref{Nuceli_class_result_table}. The suggested framework has significantly improved tissue classification, especially the connective and dead cells.
\begin{table}[h]
	\caption{Average PQ for each nuclei class accross the three datasplits of PanNuke dataset. }
	\label{Nuceli_class_result_table}
	\centering
	\footnotesize
	\begin{tabular}{p{80pt}ccccc}
	\hline
										&Neoplastic		&Inflammatory 	& Epithelium		&Dead   			&Connective	  \\  \hline
DAN-NucNet \cite{Ahmad2022}			& 0.410			& 0.314			& 0.390			& 0.000			& 0.359		\\
Micro-Net  \cite{Raza2019} 				& 0.504 			& 0.333 			& 0.442			& 0.051 			& 0.334 		  \\
DIST       \cite{Naylor2019} 				& 0.439 			& 0.343 			& 0.439			& 0.000			& 0.275 		  \\
Mask-RCNN  \cite{He2017} 				& 0.472 			& 0.290 			& 0.472			& 0.069			& 0.300 		  \\ 
HoVer-Net  \cite{Graham2019}				& 0.551 			& 0.417 			& 0.551			& 0.139 			& 0.388 		   \\
PointNu-Net 	\cite{Yao2021}				& 0.578 			& 0.433 			& 0.578			& 0.154			& 0.409		\\ \hline  
 \textbf{Proposed} 						& \textbf{0.596} 	& \textbf{0.459} 	&\textbf{0.619} 	&\textbf{0.251}	&\textbf{0.477}  \\ 
	\hline
	\end{tabular} 
\end{table}

We observe that the suggested framework produces improved semantic segmentation compared to the top performers. Since bottom-up approaches rely on semantic segmentation, we have compared the proposed framework specifically with bottom-up approaches. The comparison is made using dice metrics. The comparison is reported in Table \ref{S_I_C_Results}. The table also verifies the fact that the semantic segmentation of the bottom-up approaches improves their instance segmentation and classification, as demonstrated by \cite{Yao2021}. 
\begin{table}[!h]
	\caption{The table reports the comparison of the suggested framework with the top performer bottom-up approaches. }
	\label{S_I_C_Results}
	\centering
	\footnotesize
	\begin{tabular}{p{80pt}ccc}
	\hline
									&Semantic Segmentation	       &Instance Segmentation	&Classification	   \\ 
									&(Dice)			&(bPQ)				&(mPQ)	   		\\   
	\hline	
U-Net \cite{Ronneberger2015}			& 0.715			& \textendash 		& \textendash 	\\  
NucleiSegNet	\cite{Lal2021}			& 0.825			& \textendash 		& \textendash	\\  
DIST \cite{Naylor2019} 				& 0.717 			& 0.5346 			& 0.3406   		\\	
Micro-Net \cite{Raza2019}				& 0.813 			& 0.6053 			& 0.4059  		 \\			
HoVer-Net \cite{Graham2019}	 		& 0.819 			& 0.6596 			& 0.4629  		 \\ \hline  
\textbf{Proposed}					& \textbf{0.841}	& \textbf{0.7135} 	& \textbf{0.6337}  \\ 
	\hline
	\end{tabular} 
\end{table}

\begin{figure}[!h]
\includegraphics[width=0.95\textwidth]{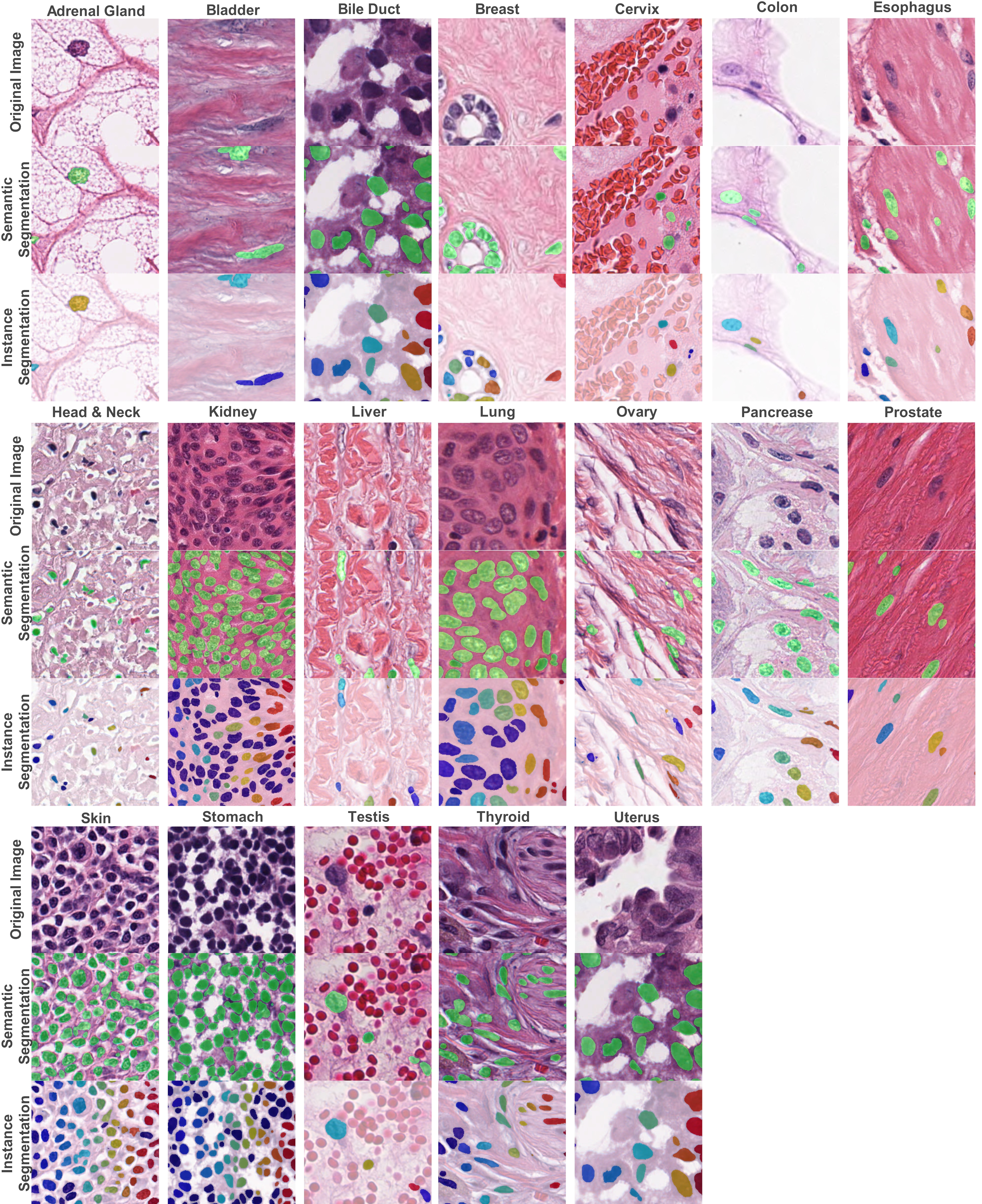}   
	\caption{The figure reports the visual results from different organs. The first, second and third row of each organ shows the original image, overlaid predicted semantic segmentation, and the overlaid predicted instances by our framework. }
	 \label{fig:Visual_Results_S_I}
\end{figure}

\begin{figure}[!h]
\centering
\includegraphics[width=0.98\textwidth]{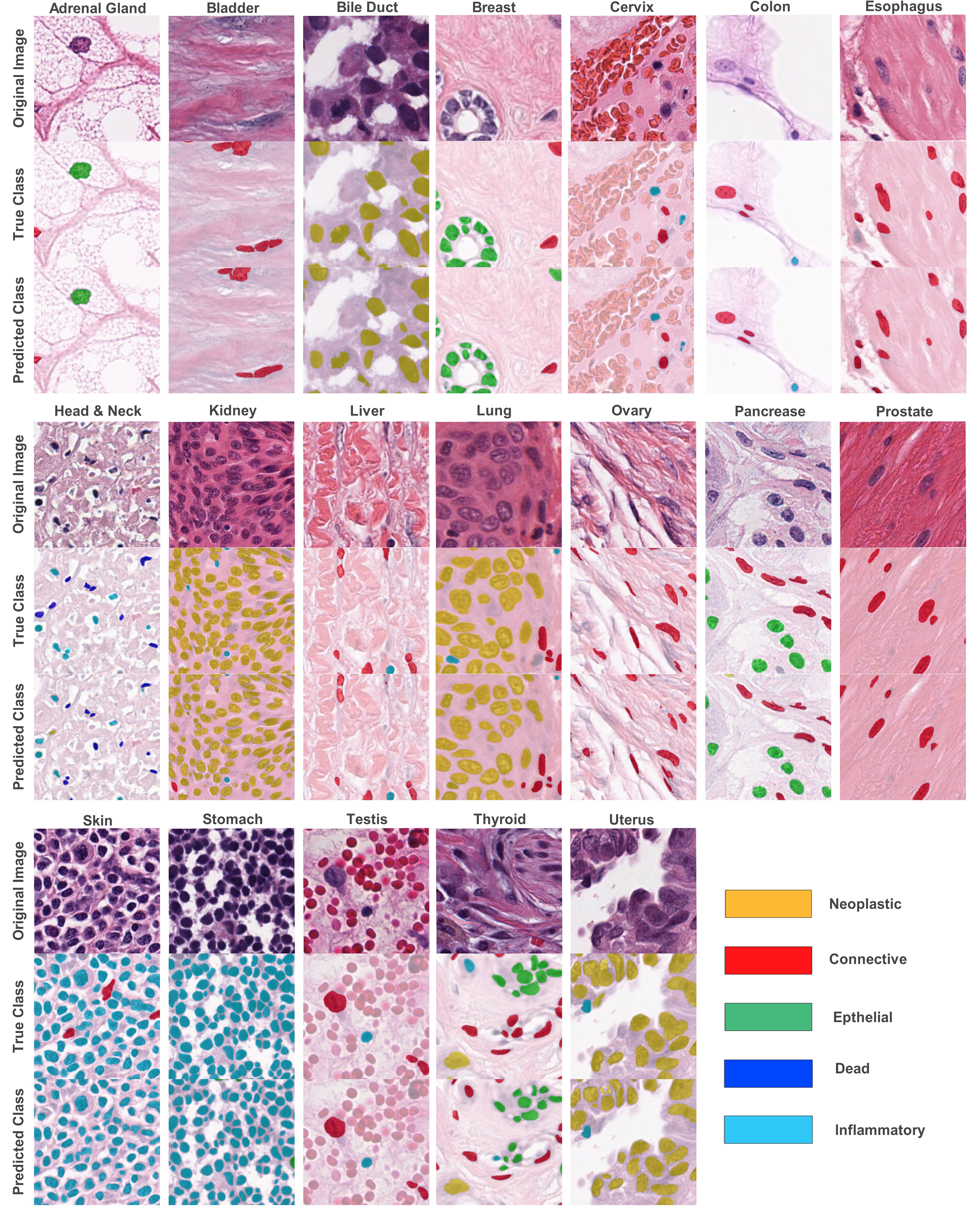}	    
	\caption{The figure reports the visual results from different organs. The first, second and third row of each organ shows the original image, overlaid true labels, and the overlaid predicted classes by our framework.}
	 \label{fig:Visual_Results_C}
\end{figure}

\begin{figure}[!h]
\centering
\includegraphics[width=0.9\columnwidth]{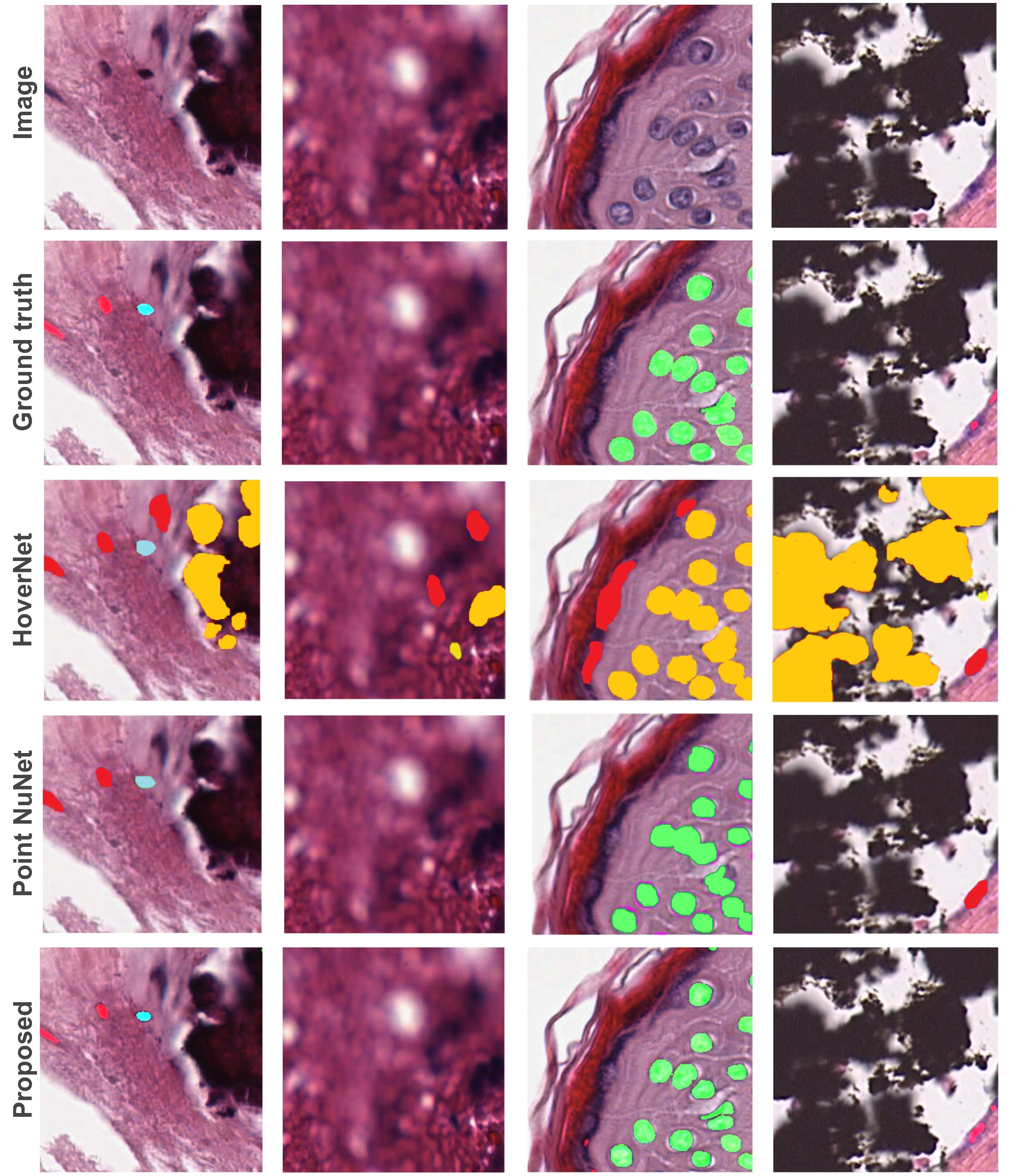}
	\caption{Example of our visual results under complex situations compared to the top performers. } \label{fig:Extreme_Comparison}
\end{figure}

\begin{figure}[!h]
\centering
\includegraphics[width=0.9\columnwidth]{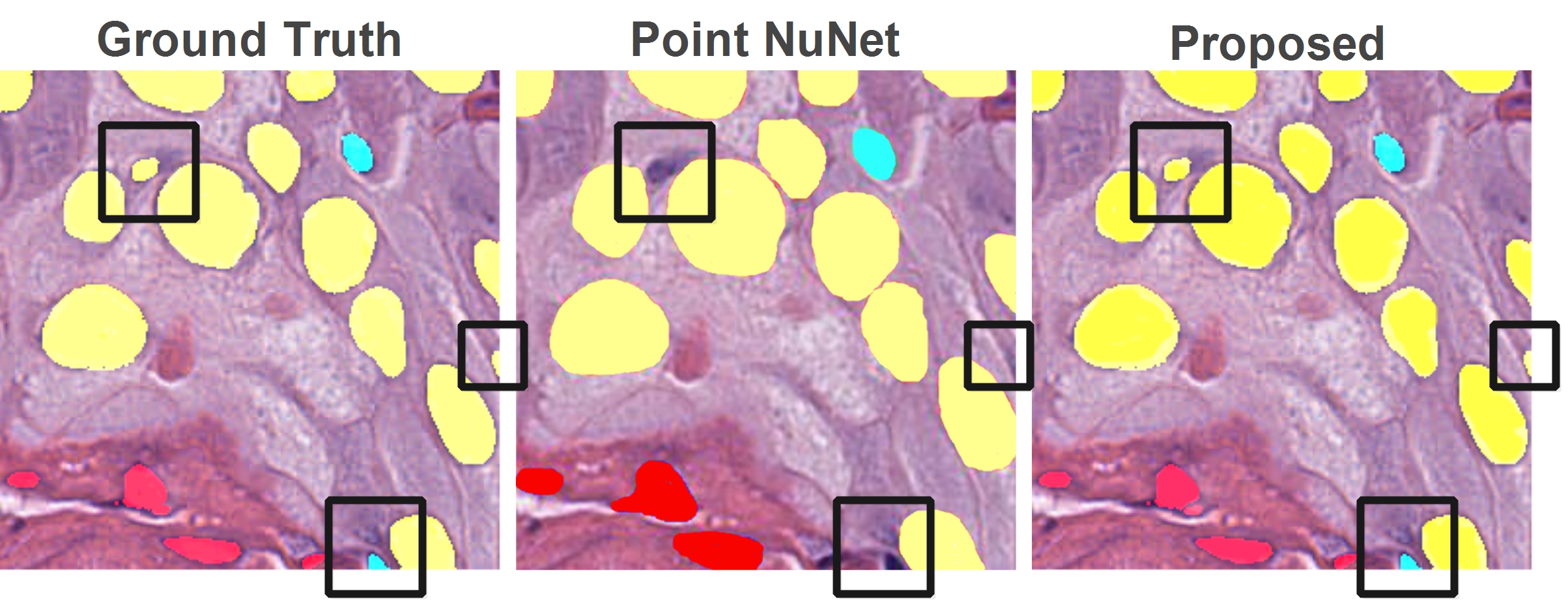}
	\caption{The visuals shows examples of the smaller nuclei, and the partial nuclei present at the edges of the image, that are being missed by the top performer. The proposed framework successfully, segmented these nuclei. } \label{fig:Left_Overs}
\end{figure}

We present visual results, under variable conditions, of our semantic and instance segmentation in Fig. \ref{fig:Visual_Results_S_I}, while classification results in Fig. \ref{fig:Visual_Results_C}. We have also presented our visual results compared to the top performers in Fig. \ref{fig:Extreme_Comparison}. The images in Fig. \ref{fig:Extreme_Comparison} also represent harsh conditions. We further compare our visual results with the top two performers, PointNu-Net \cite{Yao2021} and HoVer-Net \cite{Graham2019} (reported in Fig. \ref{fig:Left_Overs}). The suggested framework successfully segments and classifies partial or smaller nuclei. In comparison, the existing methods fail to segment smaller or partially present nuclei. 

\subsection{Significance Analysis of the Results}
\begin{figure}[!h]
\centering
\includegraphics[width=0.75\columnwidth]{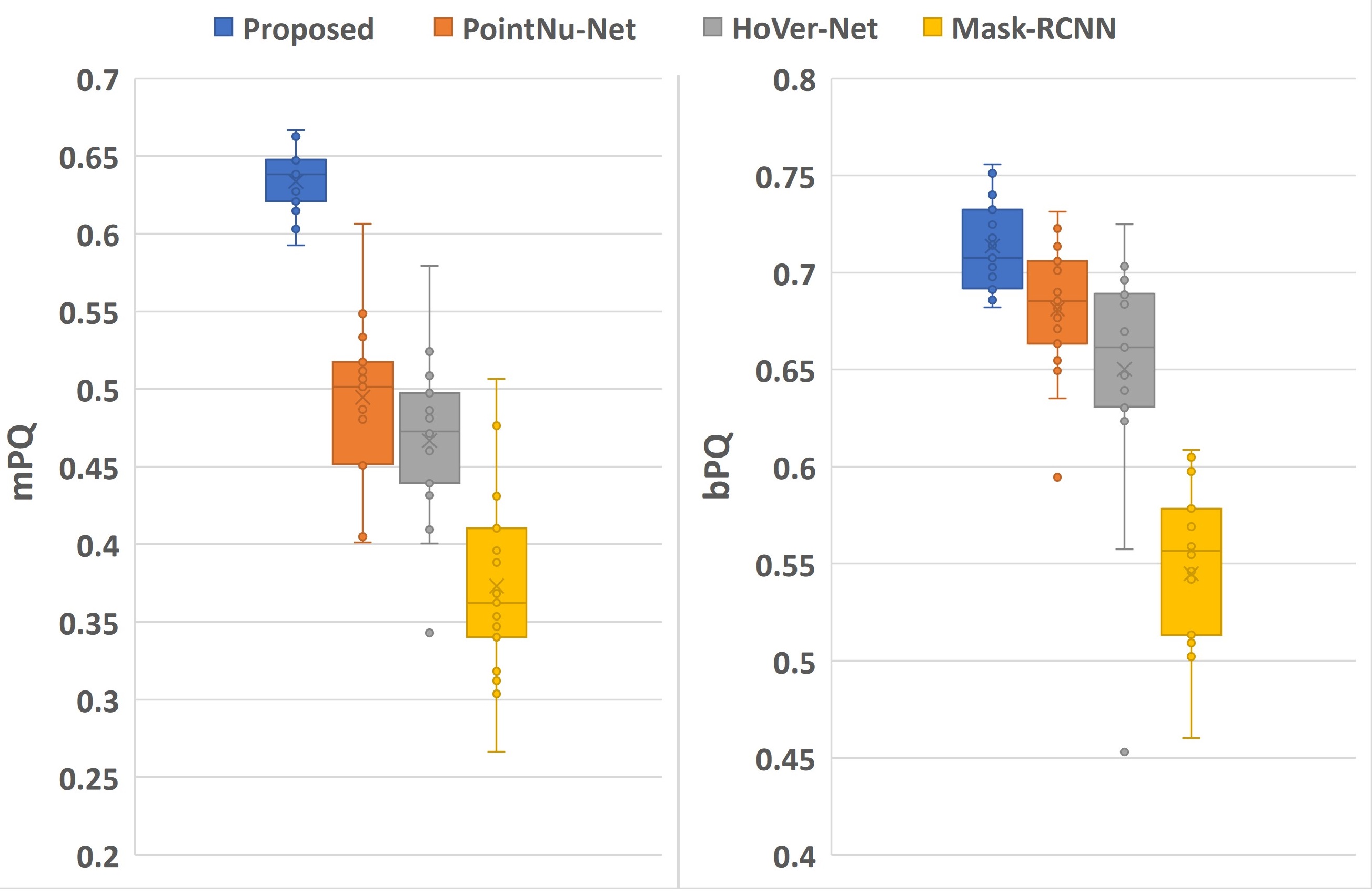}
	\caption{The box plot reports a multi-organ performance comparison of the proposed framework with the top performers. The color boxes depicts the interquartile range of the frameworks. The horizontal line within the colored rectangles shows the median performance. The horizontal peaks represent the maximum and minimum achieved performance. The dots represent the organs outliers.} \label{fig:BoxPlot}
\end{figure}

Although, the significance of the results of our framework can be observed in Table \ref{Result_Cmpr}. However, we perform an in-depth analysis to analyze our framework performance in further detail. Based on the detailed results on individual tissue sites, we plot a box plot (Fig. \ref{fig:BoxPlot}). The left boxplot in the figure is drawn, for classification, while the boxplot on the right side is drawn for instance segmentation. The boxplot helps us to analyze various aspects of the framework. 1) The color dots represent outliers among the results. We can observe that the proposed framework produces the least number of outliers, demonstrating the framework's generalization capability across multiple organs. 2) The color boxes show the interquartile range of the frameworks. The suggested framework results in a limited quartile range in comparison to the state-of-the-art, representing the framework's stability.

\section{Framework Analysis}
\subsection{Complexity Analysis of the Three-Head Model}\label{sec:Complexity_Analysis}
\begin{figure}[!h]
\centering
\includegraphics[width=\columnwidth]{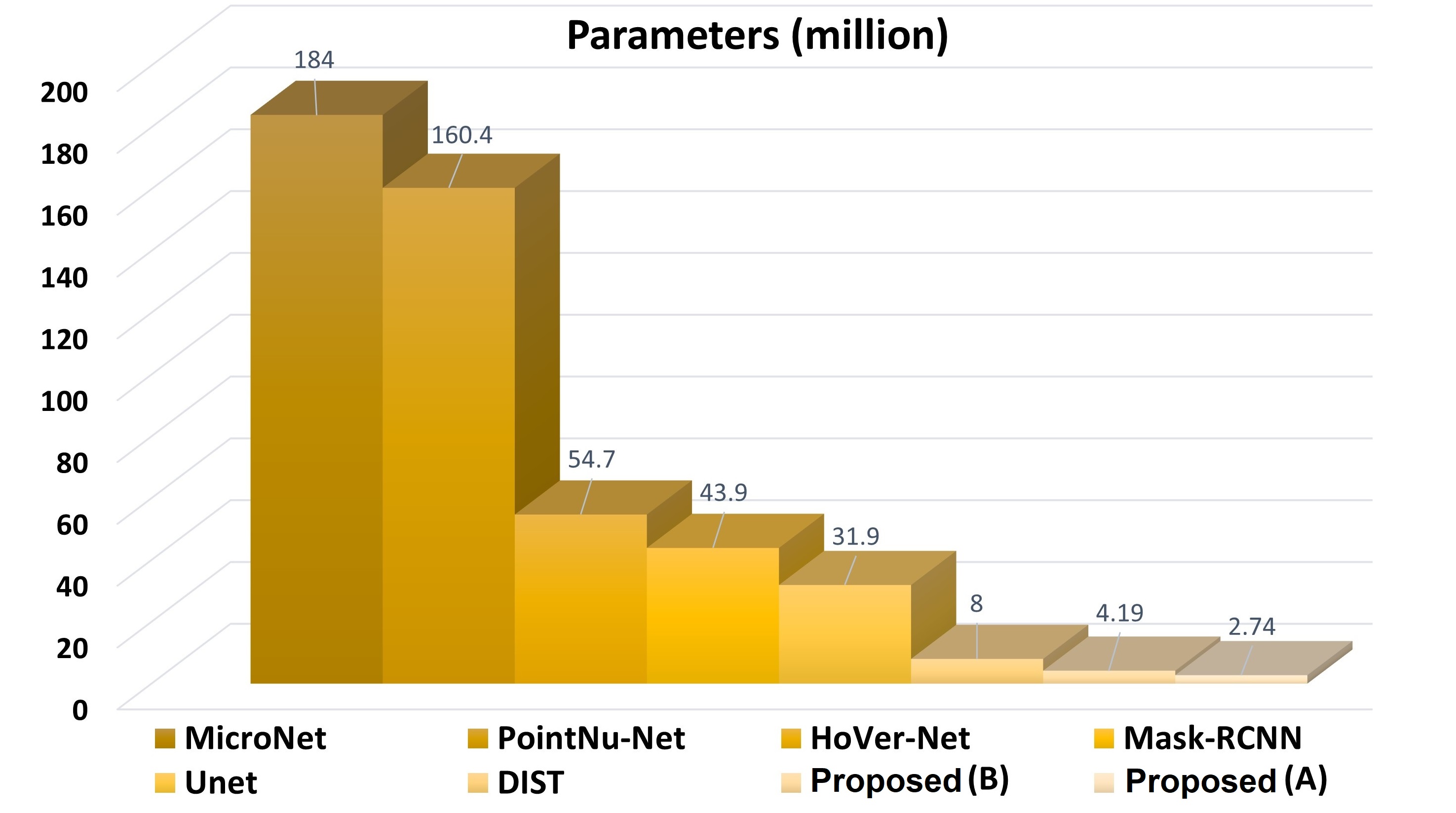}
	\caption{\textmd{Complexity comparison of the suggested three-head model with the top performing state-of-the-art in terms of trainable parameters.}}
	\label{fig:Param} 
\end{figure}

The complexity analysis of the network is an important factor in histology since a single whole slide may be of the size of 100,000x100,000 pixels. Although it is an essential comparison, it is rarely mentioned in the literature. We have compared the complexity of the suggested model with the state-of-the-art models. The comparison is in terms of the training parameters. It represents the training and inference complexity of the models. The proposed three-head model has a total of 2.74 million parameters. However, if we use a separate feature extractor for improved nuclei classification, our total parameters are increased to 4.19 million. Compared to the state-of-the-art models, the parameters are far much less. The nearest competitors are DIST \cite{Naylor2019}, and HoVer-Net \cite{Graham2019}. The figurative comparison of the parameters are shown in Fig. \ref{fig:Param}. It is to be noted that the complexity of the post-processing used by the suggested framework and the state-of-the-art are not reported. It is due to the reason that the resources utilized by the post-processing are negligible compared to the models. 

\subsection{Model Related Ablation Studies}\label{Sec:Ablation_studies_section}
For an in-depth analysis of the proposed model, we performed ablation studies related to the model's components. This study helps us to validate different aspects of the proposed method. We specifically perform four studies. First, we use a formal watershed algorithm instead of a controlled watershed. We observe that a formal watershed algorithm leads to poor instance segmentation. The use of controlled watersheds is encouraged in the previous literature compared to formal watershed methods; thus, this re-verify the previous literature. Second, we observe the framework without using pixel grouping. If we do not use pixel grouping, we get multiple class suggestions for each nuclei instance since one of our three head network heads produces pixel-wise classes. This problem is consistent with other bottom-up approaches as well. This makes the end-to-end networks for nuclei classification in histology less feasible. It also hints at the superiority of bottom-up approaches over the existing end-to-end state-of-the-art networks. Third, we use the same class weights for each nuclear category. This causes reduced performance based on the fact that the histology dataset has a high nuclei class imbalance. Because of the class imbalance, we see the worst performance of the existing methods for dead nuclei. Therefore we suggest the use of a weighted loss function for classification. Finally, we observe the framework using the same feature extractor for semantic segmentation, edge proposals, and pixel-wise classes. This has almost no effect on semantic or instance segmentation. However, using the same feature extractor reduced the classification performance. The use of a separate feature extractor is also suggested by previous works, such as Graham et al. \cite{Graham2019}. However, using a separate feature extractor increases the number of parameters. The results of the ablation studies are summarized in Table \ref{Ablation_Study}.

\begin{table}[h]
	\caption{The table reports the results obtained during the ablation studies carried out on the suggested framework.}
	\label{Ablation_Study}
	\centering
	\small
	\renewcommand{\arraystretch}{1.2}
	\begin{tabular}{p{0.5\columnwidth}cc}
	\hline
Ablation studies						& bPQ	 		& mPQ	  		\\  
	\hline
Uncontrolled Watershed 				& 0.4659		& 0.3252	 	\\
Without Pixel Grouping				& \textendash 	& 0.4786	 	\\
Equal Class Weights					& 0.7126 		& 0.5213	 	\\	
Same Feature Extractor for Class		& 0.7121	      & 0.5937	 	\\ \hline
\textbf{Proposed}					& \textbf{0.7135}& \textbf{0.6337} \\ 
	\hline
	\end{tabular}
\end{table}

\section{Conclusion}\label{sec:Conclusion}
This work presents an all-in-one framework for simultaneous semantic segmentation, instance segmentation, and nuclei classification in cancer digital histology. We use our previous DAN-NucNet model as a baseline to develop a three-head model. The three-head model has additional decoder heads that generate semantic segmentation, edge proposals, and pixel-wise class maps. We utilize the estimated edges and the semantic segmentation produced by the two decoder heads to perform instance segmentation via a controlled watershed. The edge maps help the model to improve the nuclei boundaries delineation even in challenging conditions, such as overlapping nuclei and missing staining. We then use the instance segmentation and class maps to assign classes to the nuclei instances via pixel grouping. Due to the third decoder head, pixel grouping, and weighted loss function, we observe a huge improvement in classifying imbalanced nuclear categories, i.e., connective and dead nuclei. We specifically demonstrate a considerable improvement in nuclei segmentation and classification compared to the state-of-the-art. 

Although the suggested framework has demonstrated a significant improvement in the segmentation and classification of nuclei, the classification of nuclei still has significant room for improvement. This is due to the resemblance between non-neoplastic and neoplastic nuclei. Furthermore, the high nuclei class imbalance still results in 0.261 mPQ for the dead nuclear category. In future work, these problems may be addressed.

\bibliographystyle{elsarticle-num} 
\bibliography{References}
\end{document}